\documentclass[runningheads]{llncs}

\usepackage{eccv}

\usepackage{eccvabbrv}

\usepackage{multirow}
\usepackage{graphicx}
\usepackage{booktabs}

\usepackage[accsupp]{axessibility}  

\usepackage{hyperref}

\usepackage{orcidlink}
\usepackage{marvosym}   

\begin{document}

\title{\texorpdfstring{Rethinking Garment Conditioning in Diffusion-based Virtual Try-On:\\Decouple, Don't Denoise}{Rethinking Garment Conditioning in Diffusion-based Virtual Try-On: Decouple, Don't Denoise}} 

\titlerunning{DeCo-VTON: Decouple, Don't Denoise}

\author{Kihyun Na\inst{1}\orcidlink{0009-0001-3827-5371} \and
Jinyoung Choi\inst{2}\orcidlink{0009-0002-6255-2882} \and
Injung Kim\inst{2}\orcidlink{0000-0003-4439-6097}\,\Letter}

\authorrunning{K.~Na et al.}

\institute{Advanced AI Talent Education and Research Group for Industrial Innovation, Handong Global University, Pohang 37554, Republic of Korea \\
\email{khna@handong.edu} \and
Department of Artificial Intelligence, Computer and Electrical Engineering, \\Handong Global University, Pohang 37554, Republic of Korea \\
\email{jinyoung@handong.ac.kr, ijkim@handong.edu}
}

\maketitle

\begin{abstract}
Virtual Try-On (VTON) synthesizes realistic images of a person wearing a target garment, with broad applications in e-commerce and fashion. Diffusion-based dual-UNet methods achieve strong results but double the parameters by dedicating a separate network to garment conditioning. Spatial concatenation offers a simpler single-network alternative, yet both UNet- and DiT-based instantiations report that full fine-tuning is ineffective, and the community has settled for attention-only training. We ask: why does full fine-tuning fail, and can this be resolved?
Through what is, to our knowledge, the first visualization study of dual-UNet reference network behavior, we identify a unifying insight: garment conditioning must be decoupled from the denoising process. Spatial concatenation violates this by embedding the garment within the denoising target, causing three conflicts: guidance leakage, gradient competition, and train-test discrepancy. We derive three design principles to restore this decoupling and implement them as a pure recipe atop a standard architecture with no modification. The resulting model, DeCo-VTON (860M params), achieves single-network state of the art, matching the dual-UNet state of the art at half the cost while being preferred in human evaluation.

\end{abstract}

\section{Introduction}
\label{sec:introduction}

\begin{figure}[ht]
  \centering
  \includegraphics[width=\textwidth]{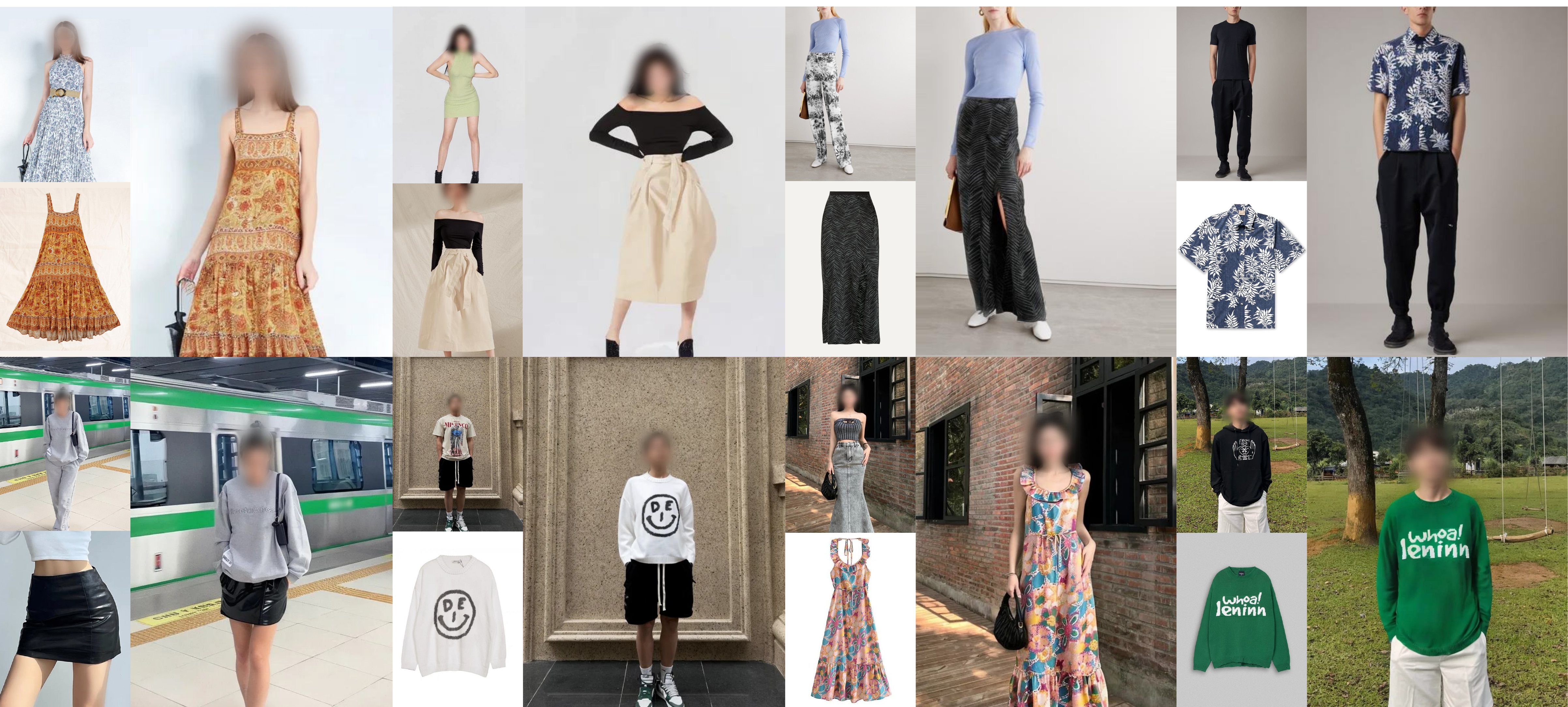}
   \caption{Virtual Try-on images generated by our DeCo-VTON on the Public datasets and In-the-Wild images. Please zoom in for more details.}
   \label{fig:first_page}
\end{figure}

Image-based virtual try-on (VTON) aims to synthesize a realistic image of a person wearing a target garment, given a person image and a garment image as input. With broad applications in fashion e-commerce---from enhancing customer experience~\cite{ecommerce1, ecommerce2, enhanced} to reducing return rates~\cite{returnrate}---VTON has attracted growing attention as a conditional image generation task. While early approaches relied on Generative Adversarial Networks~\cite{cpvton,vitonhd,pastagan,gpvton}, diffusion-based methods have become the dominant paradigm owing to their superior generation quality~\cite{ladi,dci,stableviton,idm,ootd,catvton,leffa}.

Among diffusion-based approaches, dual-UNet architectures~\cite{idm,ootd,leffa} achieve strong garment fidelity by dedicating a separate reference UNet to garment conditioning. However, this design doubles the parameter count and computational cost. Single-network alternatives have thus attracted interest for their efficiency, with \emph{spatial concatenation}---combining garment and person latents into a single denoising input---emerging as the most straightforward approach~\cite{catvton}. Yet a puzzling pattern has emerged: both CatVTON~\cite{catvton} (860M UNet) and Voost~\cite{voost} (11.9B DiT) report that full fine-tuning is ineffective under spatial concatenation, and both settle for attention-only training. Despite differing in architecture and scale by an order of magnitude, they reach the same conclusion. \textbf{Why does full fine-tuning fail, and can it be resolved?}

In this paper, we answer this question. We begin by analyzing the noise prediction behavior of dual-UNet reference networks---to our knowledge, the first such visualization study---and identify a unifying insight behind their effectiveness: \emph{the garment conditioning pathway must be decoupled from the denoising process.} Spatial concatenation violates this insight by embedding the garment latent within the denoising target. This loss of decoupling gives rise to three functional conflicts: (1)~garment information leaks into the unconditional branch, causing classifier-free guidance to suppress rather than enhance garment details; (2)~the training objective includes garment reconstruction, creating gradient competition between reconstruction and conditioning; and (3)~at inference time, the garment latent is updated by model predictions rather than the forward diffusion schedule, introducing a train-test discrepancy that accumulates over denoising steps.

We derive three design principles that restore this decoupling within the spatial-concatenation framework: Garment-Free Guidance, Decoupled Loss, and Clean Latent Anchoring, and implement them as a training and inference recipe on a standard architecture (SD1.5 inpainting UNet) with no architectural modification. The resulting model, DeCo-VTON, is the first to unlock effective full fine-tuning under spatial concatenation. With only 860M parameters, DeCo-VTON achieves state-of-the-art results among single-network methods, reaching comparable quality to the dual-UNet state of the art~\cite{leffa} at half the parameters, $2.1\times$ faster inference, and 42\% lower VRAM, while being preferred in human evaluation. Fig.~\ref{fig:first_page} shows try-on results of DeCo-VTON on the public benchmarks and on in-the-wild images.

Our contributions are as follows:
(1)~\textbf{Analysis:} We present the first visualization study of dual-UNet reference network behavior, revealing that successful garment conditioning requires decoupling from the denoising process. (2)~\textbf{Diagnosis \& Principles:} We identify three functional conflicts under spatial concatenation and derive three design principles that resolve each and unlock full fine-tuning. (3)~\textbf{Results:} Without architectural modification, DeCo-VTON achieves single-network state of the art and matches the dual-UNet state of the art at half the cost.

\section{Related Work}
\label{sec:related_work}
Early VTON models adopted GAN-based~\cite{gan, stylegan} pipelines, typically comprising a warping module for geometric alignment followed by a conditional generator for blending. VITON~\cite{viton} established this framework using Thin-Plate Spline (TPS)~\cite{tps} transformations, and subsequent methods~\cite{cpvton,vitonhd,pastagan,cvton,gpvton,acgpn} refined geometric matching through contextual cues and human parsing. While effective, these approaches depend on external modules such as pose estimators, and often produce artifacts under challenging poses or complex textures.

The advent of latent diffusion models (LDMs)~\cite{ldm}, which operate in a compact latent space learned by a Variational Auto-Encoder (VAE)~\cite{vae}, brought stronger generative priors and has since become the dominant paradigm for VTON~\cite{ladi,dci,stableviton,d4vton}. Diffusion-based VTON methods mainly differ in how they inject garment information into the denoising backbone, typically a UNet~\cite{unet} derived from Stable Diffusion~\cite{ldm} or SDXL~\cite{sdxl}. Dual-UNet architectures introduce a dedicated reference UNet that processes the garment image and provides multi-scale features to the denoising UNet. TryOnDiffusion~\cite{tryondiffusion} pioneered this design using two parallel UNets with implicit warping through cross-attention. OOTDiffusion~\cite{ootd} adopts the outfitting fusion approach, where the reference UNet features are injected via self-attention layers of the denoising UNet. IDM-VTON~\cite{idm} combines high-level garment descriptors from IP-Adapter~\cite{ipadapter} with low-level feature maps from the reference UNet. Most recently, Leffa~\cite{leffa} introduces a flow-field attention loss and additional densepose conditioning~\cite{densepose}, achieving the current dual-UNet state of the art. Despite strong fidelity, all dual-UNet methods incur substantially higher computational cost due to the additional network.

Single-network architectures remove the reference UNet to reduce overhead. LaDI-VTON~\cite{ladi} encodes garments as pseudo-word embeddings via textual inversion~\cite{textualinv}. DCI-VTON~\cite{dci} adds an explicitly warped garment~\cite{clothflow} with the masked person. CatVTON~\cite{catvton} simplifies this further through spatial concatenation of the raw garment and masked person latents, relying solely on internal self-attention without requiring additional encoders such as CLIP~\cite{clip} or DINOv2~\cite{dinov2}. More recently, Voost~\cite{voost} applies the same spatial-concatenation principle to a DiT backbone at 11.9B parameters. Notably, both CatVTON and Voost report that full fine-tuning is ineffective under spatial concatenation and instead adopt attention-only training---an observation we analyze in depth in Section~\ref{sec:analysis}.

From a broader perspective, conditioning in diffusion models ranges from cross-attention tokens to dedicated parallel networks~\cite{controlnet} to direct input-level concatenation. In the first two families the condition stays structurally separate from the denoising target---in per-modality streams of multimodal transformers~\cite{sd3} or a dedicated branch of inpainting models~\cite{brushnet}---and is thus never subject to the reconstruction loss. Spatial concatenation instead embeds the condition within the target, making the garment at once a conditioning signal and a reconstruction target; this entanglement is the implication this paper systematically studies.

\section{Analyzing Garment Conditioning}
\label{sec:analysis}

\subsection{How Dual UNets Condition on Garments}
\label{sec:dual_analysis}

\begin{figure}[ht]
  \centering
  \includegraphics[width=0.8\linewidth]{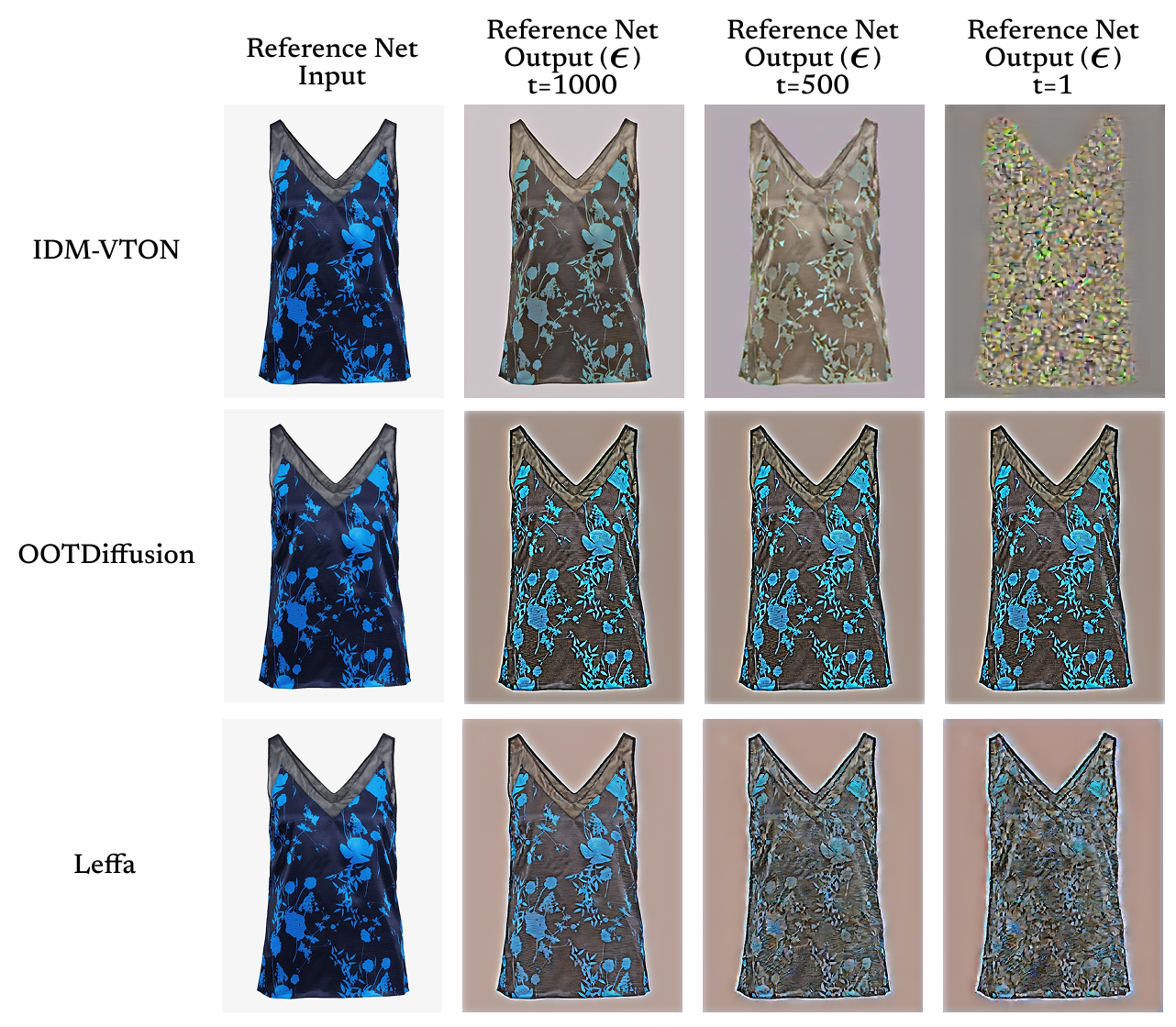}
   \caption{Visualization of the noise prediction $\boldsymbol{\epsilon}^g_t$ from the reference UNets at key timesteps. \textbf{Top:} IDM-VTON (frozen) produces strong garment features only at high~$t$, collapsing near $t{=}1$ due to the mismatch between its clean input and the pretrained noise-level expectation. \textbf{Middle:} OOTDiffusion produces strong but timestep-invariant features due to its fixed $t{=}0$ design. \textbf{Bottom:} Leffa (fine-tuned with varying~$t$) produces timestep-dependent features that transition from coarse structure to fine detail, reflecting the garment consistently across all stages.}
   \label{fig:epsilon_output}
\end{figure}

In dual-UNet VTON architectures, a dedicated reference UNet processes the garment image and provides conditioning features to the main denoising UNet.
Although these models share a common high-level design, their reference UNet strategies differ substantially: IDM-VTON~\cite{idm} freezes the pretrained weights, OOTDiffusion~\cite{ootd} fine-tunes but fixes the timestep at $t{=}0$, and Leffa~\cite{leffa} fine-tunes while feeding the same timestep~$t$ used by the denoising UNet.
These design choices lead to significant performance differences, yet no prior work has systematically analyzed why.

We briefly establish notation. In latent diffusion models, the forward process produces a noisy 
latent $\mathbf{z}_t = \sqrt{\bar{\alpha}_t}\,\mathbf{z}_0 + \sqrt{1-\bar{\alpha}_t}\,\boldsymbol{\epsilon}$ at timestep~$t$, where $\bar{\alpha}_t$ is the cumulative noise schedule and $\boldsymbol{\epsilon} \sim \mathcal{N}(\mathbf{0}, \mathbf{I})$.
A noise predictor $\boldsymbol{\epsilon}_\theta(\mathbf{z}_t, t)$ is trained to estimate~$\boldsymbol{\epsilon}$ from the noisy input, and we denote by $\mathbf{z}^g_0 = \mathcal{E}(I^g)$ the clean garment latent obtained from the VAE encoder~$\mathcal{E}$.

To understand the reference UNet's behavior, we visualize its noise prediction output $\boldsymbol{\epsilon}^g_t$ at different timesteps by passing it through the VAE decoder, as shown in Fig.~\ref{fig:epsilon_output}.
In standard diffusion models, the noise predictor $\boldsymbol{\epsilon}_\theta(\mathbf{z}_t, t)$ recovers different information depending on the timestep: primarily low-frequency, global structure at high~$t$ and high-frequency details at low~$t$~\cite{ddpm,EDM,understanding_latent, tfdsr}.
Since the reference UNet shares the same architecture, visualizing its output across timesteps reveals how it provides conditioning information to the denoising process.
To the best of our knowledge, this is the first analysis of reference UNet behavior through noise prediction visualization.

\noindent\textbf{IDM-VTON} uses a frozen pretrained SDXL UNet as the reference network.
It receives the clean garment latent $\mathbf{z}^g_0$ as input together with the denoising timestep~$t$.
However, the pretrained weights expect noisy input whose statistics match timestep~$t$: at high~$t$, the network assumes heavily corrupted input, creating a severe distributional discrepancy with the actual clean $\mathbf{z}^g_0$---a mismatch independently observed in diffusion feature extraction~\cite{cleandift}.
As a result, the extracted features become unreliable at high~$t$ (Fig.~\ref{fig:epsilon_output}, top).
At low~$t$, the discrepancy diminishes but the output exhibits low variance, limiting garment conditioning quality across the full trajectory.

\noindent\textbf{OOTDiffusion} addresses the input mismatch by fine-tuning the reference UNet for the VTON task, enabling it to extract meaningful garment features from clean input.
However, its timestep is fixed at $t{=}0$ during both training and inference (Fig.~\ref{fig:epsilon_output}, middle).
While this produces strong garment features, the output remains identical regardless of the current denoising stage---the network cannot modulate its features to match the coarse-to-fine progression of the denoising process.

\noindent\textbf{Leffa} fine-tunes the reference UNet while feeding it the same timestep~$t$ used by the denoising UNet.
This combination allows the reference UNet to learn stage-adaptive garment features: at high~$t$, it provides coarse structural information; at low~$t$, it shifts to fine-grained texture details (Fig.~\ref{fig:epsilon_output}, bottom).
This timestep-aligned conditioning is the key factor behind Leffa's superior performance among dual-UNet methods.

Comparing these three designs reveals a unifying insight: \textbf{successful garment conditioning requires the conditioning pathway to be decoupled from the denoising process.}
In dual-UNet architectures, this decoupling is structural: the reference UNet operates on clean garment input in a separate forward pass and is not subject to the denoising objective.
IDM-VTON partially benefits from this separation but fails to fully exploit it due to its frozen weights.
Leffa achieves the strongest conditioning because it combines the decoupled pathway with timestep-aligned feature generation.
As we show next, spatial concatenation fundamentally eliminates this structural decoupling.

\subsection{The Functional Mismatch in Spatial Concatenation}
\label{sec:mismatch}

Spatial concatenation combines the garment and person latents into a single input, applies noise to the entire composite, and denoises it as a whole.
This design eliminates the need for a separate reference network, offering significant parameter efficiency.
However, the garment latent becomes part of the denoising target---the structural decoupling identified in Sec.~\ref{sec:dual_analysis} is removed.

This loss of decoupling gives rise to three functional conflicts:

\noindent\textbf{Conflict~1: Garment leakage into classifier-free guidance.}
Classifier-free guidance (CFG)~\cite{cfg} amplifies the difference between conditional and unconditional predictions to strengthen the conditioning signal.
Under spatial concatenation, the unconditional prediction is generated with the garment latent still present in the input, since it is part of the denoising target rather than an external condition.
The difference between the two branches thus arises from factors other than the garment, and increasing the guidance scale~$\omega$ suppresses rather than enhances garment details.

\noindent\textbf{Conflict~2: Gradient competition between reconstruction and conditioning.}
The training objective is computed over the entire denoising target, including the garment region.
The network receives competing gradients: one directing it to accurately reconstruct the garment, and another directing it to transfer garment information to the person region.
These objectives are not aligned---features optimized for pixel-level garment reconstruction are not necessarily optimal for conditioning person generation.
In dual-UNet architectures, the reference UNet carries no reconstruction loss and serves purely as a feature extractor.

\noindent\textbf{Conflict~3: Train-test discrepancy in the garment latent.}
During training, forward diffusion provides the garment region with a correctly scheduled noisy latent at each timestep, ensuring that the noise level matches the network's expectation. At inference, however, the garment latent is updated by the model's own predictions at each step, deviating from the forward diffusion schedule. This discrepancy accumulates over steps, and the timestep-aligned conditioning that made Leffa successful (Sec.~\ref{sec:dual_analysis}) breaks down---the garment region's noise level no longer matches what the network expects at timestep~$t$.

These conflicts manifest at two observable levels.
Conflicts~2 and~3 degrade the conditional branch itself: even without CFG ($\omega{=}1.0$), garment details appear blurred because the network's garment encoding capacity is compromised.
Conflict~1 causes further degradation when CFG is applied: increasing~$\omega$ amplifies garment suppression rather than enhancement.
We provide quantitative evidence for both effects in Sec.~\ref{sec:ablation} and ~\ref{sec:cfg_analysis}.

Cross-architecture evidence supports this analysis. CatVTON~\cite{catvton} (860M UNet) reports that full fine-tuning yields no improvement over attention-only training. Voost~\cite{voost} (11.9B DiT) reports that full fine-tuning (FID~6.351) underperforms attention-only training (FID~5.269) with less than a quarter of the trainable parameters. The same conclusion emerging from fundamentally different architectures and scales strongly suggests that the problem lies in spatial concatenation itself, not in any particular backbone.

We argue that the ineffectiveness of full fine-tuning observed in previous studies stems from these unresolved conflicts rather than from an intrinsic limitation of full fine-tuning itself. If the three conflicts can be explicitly resolved within the spatial-concatenation framework, full fine-tuning should unlock stronger garment conditioning by allowing the entire network to adapt to the try-on task.
We derive three design principles---each resolving one conflict---and present their implementations in the following section.

\section{Decoupled Conditioning for Virtual Try-On}
\label{sec:method}

We instantiate three design principles---each resolving one conflict from Sec.~\ref{sec:mismatch}---as a training and inference recipe applied to CatVTON~\cite{catvton}, deliberately keeping the architecture identical: same backbone (SD1.5 inpainting UNet, 860M parameters), same input formulation, same optimizer and learning rate.

\subsection{Overview}
\label{sec:overview}

\begin{figure}[ht]
    \centering
    \includegraphics[width=\linewidth]{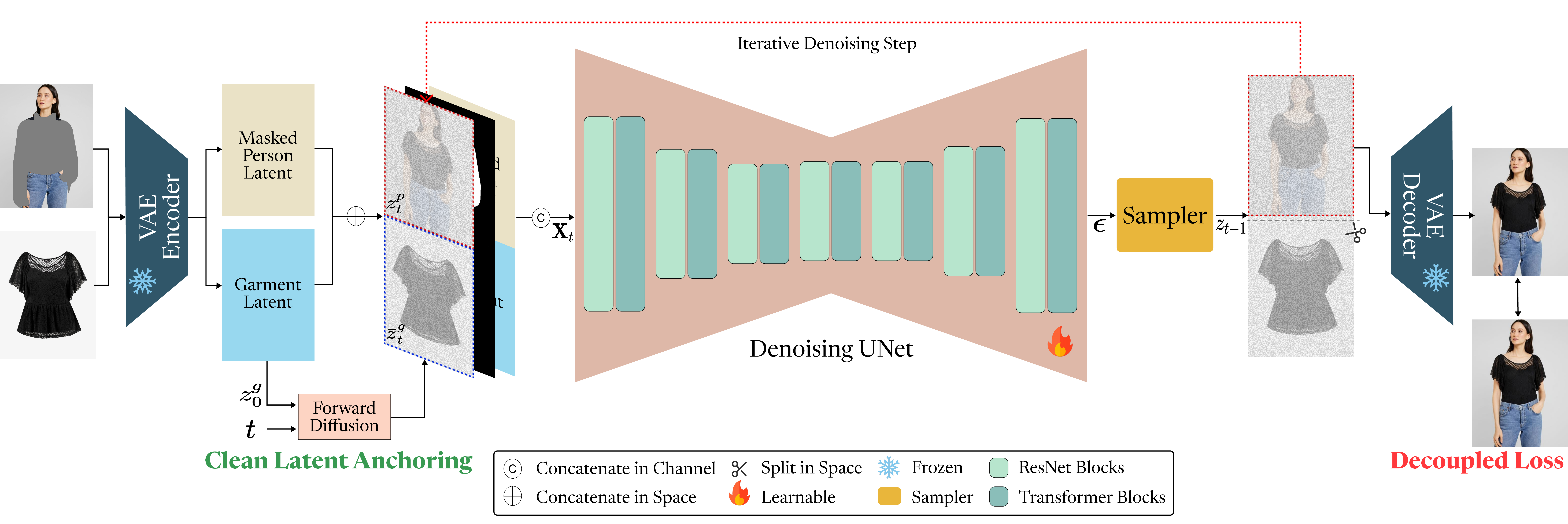}
    \caption{\textbf{Overview of DeCo-VTON}. The masked person image and the garment image are encoded into latent representations using a VAE and spatially concatenated to form the denoising input. The Decoupled Loss supervises only the person region, allowing the garment region to serve purely as a conditioning signal. At each inference step, Clean Latent Anchoring replaces the garment latent with the forward-diffused ground truth, ensuring timestep-aligned conditioning throughout the denoising trajectory. The final latent is decoded to produce the try-on output.}
    \label{fig:architecture}
\end{figure}

CatVTON conditions the denoising process through spatial concatenation of garment and person latents.
Given a masked person image $I^p$ and a garment image $I^g$, the VAE encoder produces latents $\mathbf{z}^p_0 = \mathcal{E}(I^p)$ and $\mathbf{z}^g_0 = \mathcal{E}(I^g)$.
These are spatially concatenated along the height dimension, and the input to the denoising UNet at timestep $t$ is:
\begin{equation}
    \mathbf{X}_t = \operatorname{Concat_{ch}}\!\left(\mathbf{z}^p_t \oplus \mathbf{z}^g_t,\; \mathbf{M} \oplus \mathbf{0},\; \mathbf{z}^p_0 \oplus \mathbf{z}^g_0\right),
    \label{eq:catvton_input}
\end{equation}
where $\oplus$ denotes spatial concatenation along the height dimension, $\operatorname{Concat_{ch}}$ denotes channel-wise concatenation, and $\mathbf{M}$ is the inpainting mask.
The standard training objective computes the noise prediction loss over the entire spatially concatenated output:
\begin{equation}
    \mathcal{L}_{\text{standard}} = \mathbb{E}_{\mathbf{z}_0, \boldsymbol{\epsilon}, t}\!\left[\left\|\boldsymbol{\epsilon} - \boldsymbol{\epsilon}_\theta(\mathbf{X}_t, t)\right\|^2\right],
    \label{eq:standard_loss}
\end{equation}
where $\boldsymbol{\epsilon} = \boldsymbol{\epsilon}^p \oplus \boldsymbol{\epsilon}^g$ includes noise from both regions.
As analyzed in Sec.~\ref{sec:mismatch}, this formulation---combined with the standard CFG and inference procedure---gives rise to three conflicts that prevent full fine-tuning from succeeding.

DeCo-VTON applies full fine-tuning to the entire UNet and introduces a training and inference recipe comprising three principles (Fig.~\ref{fig:architecture}):
Garment-Free Guidance (Sec.~\ref{sec:gfg}), Decoupled Loss (Sec.~\ref{sec:dl}), and Clean Latent Anchoring (Sec.~\ref{sec:cla}).

\subsection{Garment-Free Guidance}
\label{sec:gfg}

Conflict~1 (Sec.~\ref{sec:mismatch}) arises because the unconditional branch in standard CFG retains garment information.
In dual-UNet methods, disabling the reference network naturally yields a garment-free unconditional prediction.
Spatial concatenation requires an explicit mechanism to achieve the same effect.

In CatVTON's CFG, the unconditional input removes the clean garment latent $\mathbf{z}^g_0$ from the conditioning channels but retains the noisy garment latent $\mathbf{z}^g_t$:
\begin{equation}
    \mathbf{X}^{\text{uncond}}_t = \operatorname{Concat_{ch}}\!\left(\mathbf{z}^p_t \oplus {\mathbf{z}^g_t},\; \mathbf{M} \oplus \mathbf{0},\; \mathbf{z}^p_0 \oplus \mathbf{0}\right).
    \label{eq:cfg_catvton}
\end{equation}
The retained ${\mathbf{z}^g_t}$ leaks garment information into the unconditional prediction, causing CFG to suppress garment details as the guidance scale~$\omega$ increases.

We construct a strictly garment-free unconditional branch by removing all garment-related latents:
\begin{equation}
    \mathbf{X}^{\text{gf}}_t = \operatorname{Concat_{ch}}\!\left(\mathbf{z}^p_t \oplus {\mathbf{0}},\; \mathbf{M} \oplus \mathbf{0},\; \mathbf{z}^p_0 \oplus \mathbf{0}\right).
    \label{eq:cfg_ours}
\end{equation}
The guided prediction then becomes:
\begin{equation}
    \hat{\boldsymbol{\epsilon}}_t = \boldsymbol{\epsilon}_\theta(\mathbf{X}^{\text{gf}}_t, t) + \omega\!\left(\boldsymbol{\epsilon}_\theta(\mathbf{X}_t, t) - \boldsymbol{\epsilon}_\theta(\mathbf{X}^{\text{gf}}_t, t)\right)
    \label{eq:gfg}
\end{equation}
By making the unconditional branch entirely garment-free, the guidance signal, \ie the difference between the two predictions, now correctly isolates the garment's contribution; the conditional branch retains full garment information, so the guidance amplifies rather than suppresses garment details. We term this scheme Garment-Free Guidance (GFG), where \emph{garment-free} refers specifically to the unconditional branch, analogous to how disabling the reference network in a dual-UNet model yields an unconditional prediction that serves as a clean person prior without any garment information. During training, we randomly drop the garment condition with probability $10\%$ to enable the model to learn the garment-free prediction.
\subsection{Decoupled Loss}
\label{sec:dl}

Conflict~2 (Sec.~\ref{sec:mismatch}) arises because the training objective in Eq.~\ref{eq:standard_loss} includes the garment region, creating gradient competition between garment reconstruction and person generation. In dual-UNet methods, the reference UNet carries no reconstruction loss and serves purely as a feature extractor. To replicate this separation, we exclude the garment region from the training objective.

Since the person and garment latents are spatially concatenated along the height dimension (person on top, garment on bottom), the full UNet output naturally decomposes as $\boldsymbol{\epsilon}_\theta = \boldsymbol{\epsilon}^p_\theta \oplus \boldsymbol{\epsilon}^g_\theta$, where $\boldsymbol{\epsilon}^p_\theta$ denotes the person-region portion. We compute the loss only on the person region:
\begin{equation}
    \mathcal{L}_{\text{decoupled}} = \mathbb{E}_{\mathbf{z}_0, \boldsymbol{\epsilon}, t}\!\left[\left\| \bar{\boldsymbol{\epsilon}}^p_{\text{dream},t} - \boldsymbol{\epsilon}^p_\theta(\tilde{\mathbf{X}}_t, t)\right\|^2\right],
    \label{eq:decoupled_loss}
\end{equation}
where we adopt the DREAM~\cite{dream} objective with its target rectification and input rectification applied exclusively to the person region, keeping the garment region fixed. The rectified input $\tilde{\mathbf{X}}_t$ is constructed by substituting the DREAM-corrected person latent into Eq.~\ref{eq:catvton_input} while leaving the garment latent unchanged. 
No mask generation or pixel-wise weighting is required---the spatial layout directly enables the decomposition. Full details are provided in the supplementary material.

\subsection{Clean Latent Anchoring}
\label{sec:cla}

Conflict~3 (Sec.~\ref{sec:mismatch}) arises because the garment latent at inference is updated by the model's own predictions, deviating from the forward diffusion schedule and breaking the timestep-aligned conditioning that made Leffa successful in the dual-UNet setting (Sec.~\ref{sec:dual_analysis}).

We distinguish $\bar{\mathbf{z}}^g_t$ (computed from $\mathbf{z}^g_0$ via forward diffusion) from $\mathbf{z}^g_t$ (updated by the model's prediction chain during standard inference). Since the clean garment latent $\mathbf{z}^g_0$ is available as input, we compute the ground-truth noisy latent at any timestep via forward diffusion:
\begin{equation}
    \bar{\mathbf{z}}^g_t = \sqrt{\bar{\alpha}_t}\,\mathbf{z}^g_0 + \sqrt{1-\bar{\alpha}_t}\,\boldsymbol{\epsilon}_{\text{init}}, \quad \boldsymbol{\epsilon}_{\text{init}} \sim \mathcal{N}(\mathbf{0}, \mathbf{I}).
    \label{eq:cla}
\end{equation}
At each denoising step, we replace $\mathbf{z}^g_t$ with $\bar{\mathbf{z}}^g_t$ in Eq.~\ref{eq:catvton_input}. A fixed noise sample $\boldsymbol{\epsilon}_{\text{init}}$ is used  throughout the denoising process to maintain consistency. This ensures two properties: (1)~no error accumulation, since the garment latent is always derived directly from $\mathbf{z}^g_0$  rather than from the prediction chain; and (2)~exact timestep alignment, since the noise level in the garment region matches what the network expects at timestep~$t$.

CLA is an inference-only modification. During training, forward diffusion already provides the garment region with correctly scheduled noisy latents, so the training procedure requires no change. This alignment between training and inference distributions is precisely what CLA restores. We note that conditioning the sampling process on a known signal has been explored in diffusion-based inpainting~\cite{repaint} and noising-denoising editing~\cite{sdedit}. CLA adopts a similar known-signal anchoring during sampling but serves a distinct purpose: rather than inpainting unknown pixels, it maintains timestep-aligned garment conditioning within the spatial-concatenation framework (Sec.~\ref{sec:dual_analysis}), preventing error accumulation in the conditioning (not target) region.

\section{Experiments}
\label{sec:experiments}

\subsection{Setup}
\label{sec:setup}

\noindent\textbf{Datasets.}
We evaluate on two public benchmarks: VITON-HD~\cite{vitonhd} (11,647 training / 2,032 testing upper-body pairs) and DressCode~\cite{dresscode} (48,392 training / 5,400 testing full-body pairs across tops, bottoms, and dresses).
For DressCode, garment-agnostic masks are generated using DensePose~\cite{densepose} and the SCHP parser~\cite{atr,lip,schp}.

\noindent\textbf{Implementation.}
We adopt the inpainting variant of Stable Diffusion 1.5~\cite{ldm} as the backbone---identical to CatVTON~\cite{catvton}.
Training is conducted at $512{\times}384$ resolution using AdamW~\cite{adamw} with batch size 128, learning rate $1{\times}10^{-5}$, and garment dropout probability of $0.1$.
We train for 16K steps on VITON-HD and 32K steps on DressCode on 2 NVIDIA H200 GPUs (approximately 10 and 20 hours, respectively).
The guidance scale is set to $\omega{=}2.5$ following~\cite{catvton,ootd}.

\noindent\textbf{Metrics.}
We report FID~\cite{fid, cleanfid}, KID~\cite{kid}, SSIM~\cite{ssim}, and LPIPS~\cite{lpips} for paired evaluation, and FID and KID for unpaired evaluation.

\subsection{Quantitative Comparison}
\label{sec:quantitative}

\begin{table}[t!]
  \centering
  \footnotesize
  \setlength{\tabcolsep}{4pt}
  \caption{Quantitative comparison on VITON-HD. \textbf{Best} in bold, \underline{second-best} underlined. All methods are UNet/CNN-based; for DiT-based comparisons, see Appendix~B.}
  \label{tab:vitonhd}
  \begin{tabular}{l cccc cc}
    \toprule
    \multirow{2}{*}{Method} & \multicolumn{4}{c}{Paired} & \multicolumn{2}{c}{Unpaired} \\
    \cmidrule(lr){2-5} \cmidrule(lr){6-7}
    & FID$\downarrow$ & KID$\downarrow$ & SSIM$\uparrow$ & LPIPS$\downarrow$ & FID$\downarrow$ & KID$\downarrow$ \\
    \midrule
    HR-VTON~\cite{hrvton}        & 10.88 & 4.480 & 0.876 & 0.097 & 13.06 & 4.720 \\
    GP-VTON~\cite{gpvton}        &  8.726 & 3.944 & 0.870 & 0.059 & 11.84 & 4.310 \\
    DCI-VTON~\cite{dci}          &  9.408 & 4.547 & 0.862 & 0.061 & 12.53 & 5.251 \\
    LaDI-VTON~\cite{ladi}        &  6.660 & 1.080 & 0.876 & 0.091 &  9.410 & 1.600 \\
    StableVITON~\cite{stableviton} & 6.439 & 0.942 & 0.854 & 0.091 & 11.05 & 3.914 \\
    IDM-VTON~\cite{idm}          &  5.762 & 0.732 & 0.850 & 0.060 & 14.65 & 8.754 \\
    PromptDresser~\cite{promptdressor} & 8.540 & 0.670 & 0.869 & 0.112 & -- & -- \\
    OOTDiffusion~\cite{ootd}     &  9.305 & 4.086 & 0.819 & 0.088 & 12.41 & 4.689 \\
    CatVTON~\cite{catvton}       &  5.425 & 0.411 & 0.870 & 0.057 &  9.015 & 1.091 \\
    Leffa~\cite{leffa}           & \underline{4.540} & \underline{0.050} & \textbf{0.899} & \underline{0.048} & \underline{8.520} & \textbf{0.320} \\
    \textbf{DeCo-VTON (Ours)}    & \textbf{4.438} & \textbf{0.010} & \underline{0.880} & \textbf{0.047} & \textbf{8.266} & \underline{0.517} \\
    \bottomrule
  \end{tabular}
\end{table}

\begin{table}[t!]
  \centering
  \footnotesize
  \setlength{\tabcolsep}{4pt}
  \caption{Quantitative comparison on DressCode. Notation follows Table~\ref{tab:vitonhd}.}
  \label{tab:dresscode}
  \begin{tabular}{l cccc cc}
    \toprule
    \multirow{2}{*}{Method} & \multicolumn{4}{c}{Paired} & \multicolumn{2}{c}{Unpaired} \\
    \cmidrule(lr){2-5} \cmidrule(lr){6-7}
    & FID$\downarrow$ & KID$\downarrow$ & SSIM$\uparrow$ & LPIPS$\downarrow$ & FID$\downarrow$ & KID$\downarrow$ \\
    \midrule
    GP-VTON~\cite{gpvton}        &  9.927 & 4.610 & 0.771 & 0.180 & 12.79 & 6.627 \\
    LaDI-VTON~\cite{ladi}        &  9.555 & 4.683 & 0.766 & 0.237 & 10.68 & 5.787 \\
    IDM-VTON~\cite{idm}          &  6.821 & 2.924 & 0.880 & 0.056 &  9.546 & 4.320 \\
    OOTDiffusion~\cite{ootd}     &  4.610 & 0.955 & 0.885 & 0.053 & 12.57 & 6.627 \\
    CatVTON~\cite{catvton}       &  3.992 & 0.818 & 0.892 & \underline{0.046} &  6.137 & 1.403 \\
    Leffa~\cite{leffa}           & \textbf{2.060} & \underline{0.070} & \textbf{0.924} & \textbf{0.031} & \underline{4.480} & \textbf{0.620} \\
    \textbf{DeCo-VTON (Ours)}    & \underline{2.175} & \textbf{0.062} & \underline{0.914} & \textbf{0.031} & \textbf{4.310} & \underline{0.628} \\
    \bottomrule
  \end{tabular}
\end{table}

Tables~\ref{tab:vitonhd} and~\ref{tab:dresscode} compare DeCo-VTON against UNet-based methods on VITON-HD and DressCode.
On VITON-HD, DeCo-VTON achieves the best FID, KID, and LPIPS in the paired setting, trailing Leffa only in SSIM.
On DressCode, DeCo-VTON matches Leffa in LPIPS and yields a lower KID, while achieving the best unpaired FID.
Notably, DeCo-VTON uses a single 860M UNet, whereas Leffa requires a dual-UNet architecture with 1.8B parameters.
These results demonstrate that resolving the functional mismatch through our conditioning recipe is sufficient to close the single--dual network performance gap without architectural changes. Results at higher resolution ($1024{\times}768$) are reported in Appendix~B, where DeCo-VTON likewise attains the best paired FID on both benchmarks, confirming that the recipe remains effective at high resolution.

\subsection{Efficiency Analysis}
\label{sec:efficiency}

\begin{table}[t]
  \centering
  \footnotesize
  \setlength{\tabcolsep}{4pt}
  \caption{Efficiency comparison at $512{\times}384$ resolution. Latency and memory are measured on a single H200 GPU with FP16.}
  \label{tab:efficiency}
  \begin{tabular}{l cccc}
    \toprule
    Method & Params (M) & GFLOPs & Latency (s) & VRAM (GB) \\
    \midrule
    OOTDiffusion~\cite{ootd} & 2229.73 & 1225.16 & 1.5 & 5.93 \\
    IDM-VTON~\cite{idm}      & 7003.26 & 2679.45 & 6.6 & 14.62 \\
    CatVTON~\cite{catvton}   & 859.54 &  973.99 & 1.3 & 2.26 \\
    Leffa~\cite{leffa}       & 1802.72 & 1012.03 & 2.7 & 3.91 \\
    \textbf{DeCo-VTON (Ours)} & \textbf{859.54} & \textbf{973.99} & \textbf{1.3} & \textbf{2.26} \\
    \bottomrule
  \end{tabular}
\end{table}

DeCo-VTON shares the identical backbone with CatVTON, so its computational cost is unchanged: 860M parameters, 974 GFLOPs, 1.3s latency, and 2.26 GB peak VRAM (Table~\ref{tab:efficiency}).
Compared to Leffa, DeCo-VTON achieves comparable quality at $2.1{\times}$ faster inference and $42\%$ lower memory, confirming that a properly designed conditioning recipe can match dual-UNet performance without the associated overhead. Measurement details are provided in the supplementary material.

\subsection{Qualitative Comparison}
\label{sec:qualitative}

\begin{figure}[ht!]
  \centering
  \includegraphics[height=0.7\textheight]{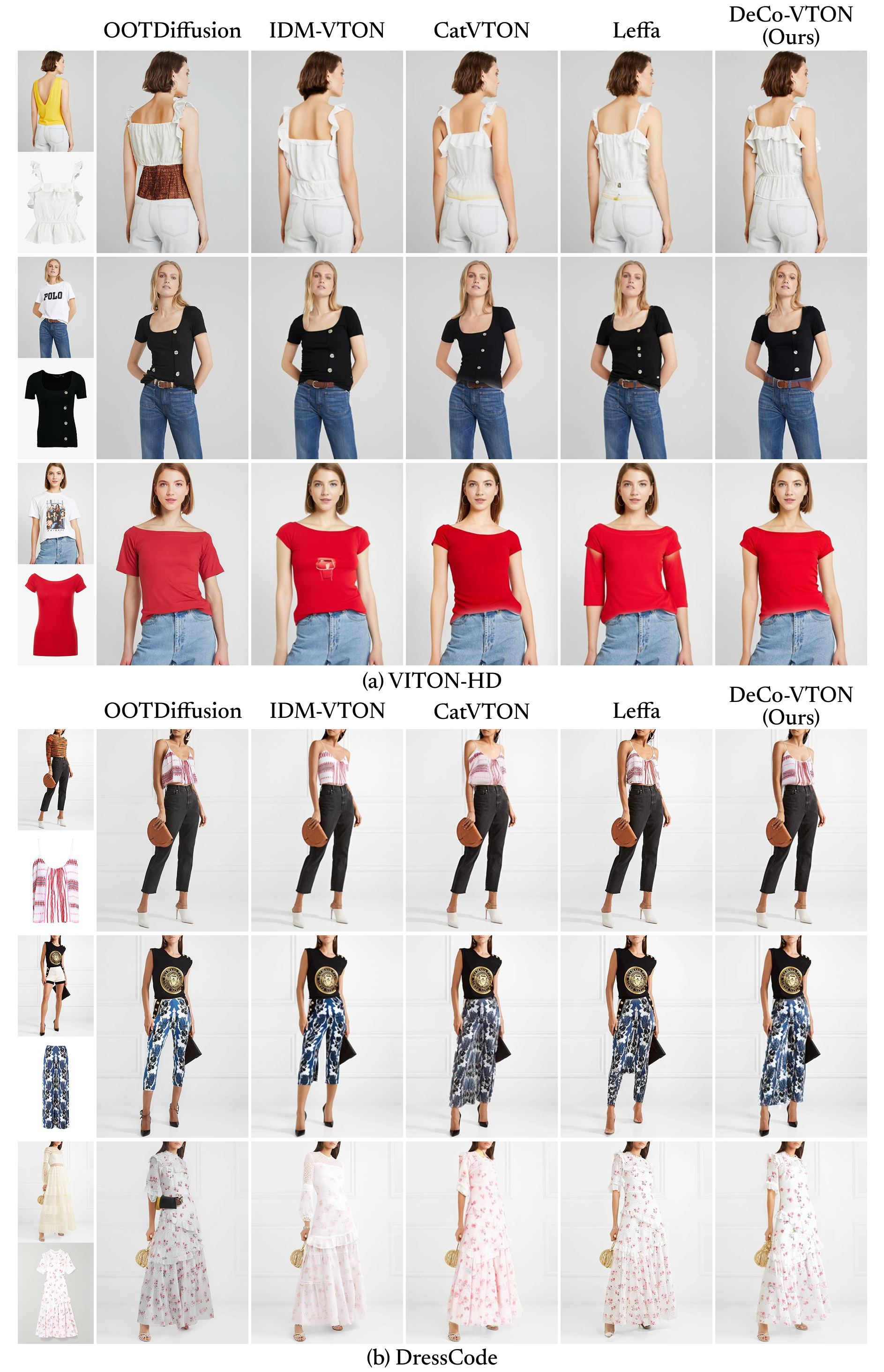}
  \caption{Qualitative comparison on VITON-HD and DressCode. DeCo-VTON preserves garment details (logos, text, patterns) more faithfully than CatVTON and produces results comparable to the dual-UNet model Leffa, despite using a single UNet with half the parameters.}
  \label{fig:qualitative}
\end{figure}

Fig.~\ref{fig:qualitative} shows qualitative results on both datasets.
DeCo-VTON preserves fine garment details---logos, textures, and complex patterns---more faithfully than CatVTON, and produces results visually comparable to Leffa.

\subsection{Ablation Study}
\label{sec:ablation}

\begin{table}[t]
  \centering
  \footnotesize
  \setlength{\tabcolsep}{4pt}
  \caption{Progressive ablation on VITON-HD. GFG: Garment-Free Guidance, DL: Decoupled Loss, CLA: Clean Latent Anchoring. CatVTON results are from the original paper~\cite{catvton}. Each row adds one principle to the previous configuration.}
  \label{tab:ablation}
  \begin{tabular}{l cccc cc}
    \toprule
    \multirow{2}{*}{Method} & \multicolumn{4}{c}{Paired} & \multicolumn{2}{c}{Unpaired} \\
    \cmidrule(lr){2-5} \cmidrule(lr){6-7}
    & FID$\downarrow$ & KID$\downarrow$ & SSIM$\uparrow$ & LPIPS$\downarrow$ & FID$\downarrow$ & KID$\downarrow$ \\
    \midrule
    CatVTON (Attn-only) & 5.425 & 0.411 & 0.870 & 0.057 & 9.015 & 1.091 \\
    CatVTON (Full FT)   & 5.250 & 0.402 & 0.869 & 0.055 & 8.813 & 0.956 \\
    \midrule
    + GFG               & 4.538 & 0.074 & 0.880 & 0.048 & 8.374 & 0.610 \\
    + DL     & 4.517 & 0.068 & 0.880 & 0.048 & 8.338 & 0.596 \\
    + CLA (= DeCo-VTON)  & \textbf{4.438} & \textbf{0.010} & \textbf{0.880} & \textbf{0.047} & \textbf{8.266} & \textbf{0.517} \\
    \bottomrule
  \end{tabular}
\end{table}

Table~\ref{tab:ablation} validates each principle through progressive addition.
Two observations directly support the analysis in Sec.~\ref{sec:mismatch}.

First, CatVTON with full fine-tuning (row~2) yields only marginal improvement over attention-only training (row~1), consistent with the functional mismatch preventing full FT from reaching its potential.

Second, resolving Conflict~1 via GFG (row~3) yields the largest single gain (FID: $5.250 {\to} 4.538$, KID: $0.402 {\to} 0.074$), confirming that garment leakage into the unconditional branch is the most severe bottleneck under spatial concatenation.
Decoupled Loss (row~4) and CLA (row~5) provide further gains by eliminating gradient competition and restoring timestep alignment, respectively, with CLA yielding a notable KID reduction ($0.068 {\to} 0.010$).

We further isolate DL's individual contribution with a leave-one-out comparison at both resolutions in Appendix~D, where its effect is robust to the ablation ordering at $512{\times}384$ and becomes more pronounced at $1024{\times}768$.

\subsection{Diagnostic Analysis: CFG Scale}
\label{sec:cfg_analysis}

\begin{figure}[t]
  \centering
  \includegraphics[width=0.7\linewidth]{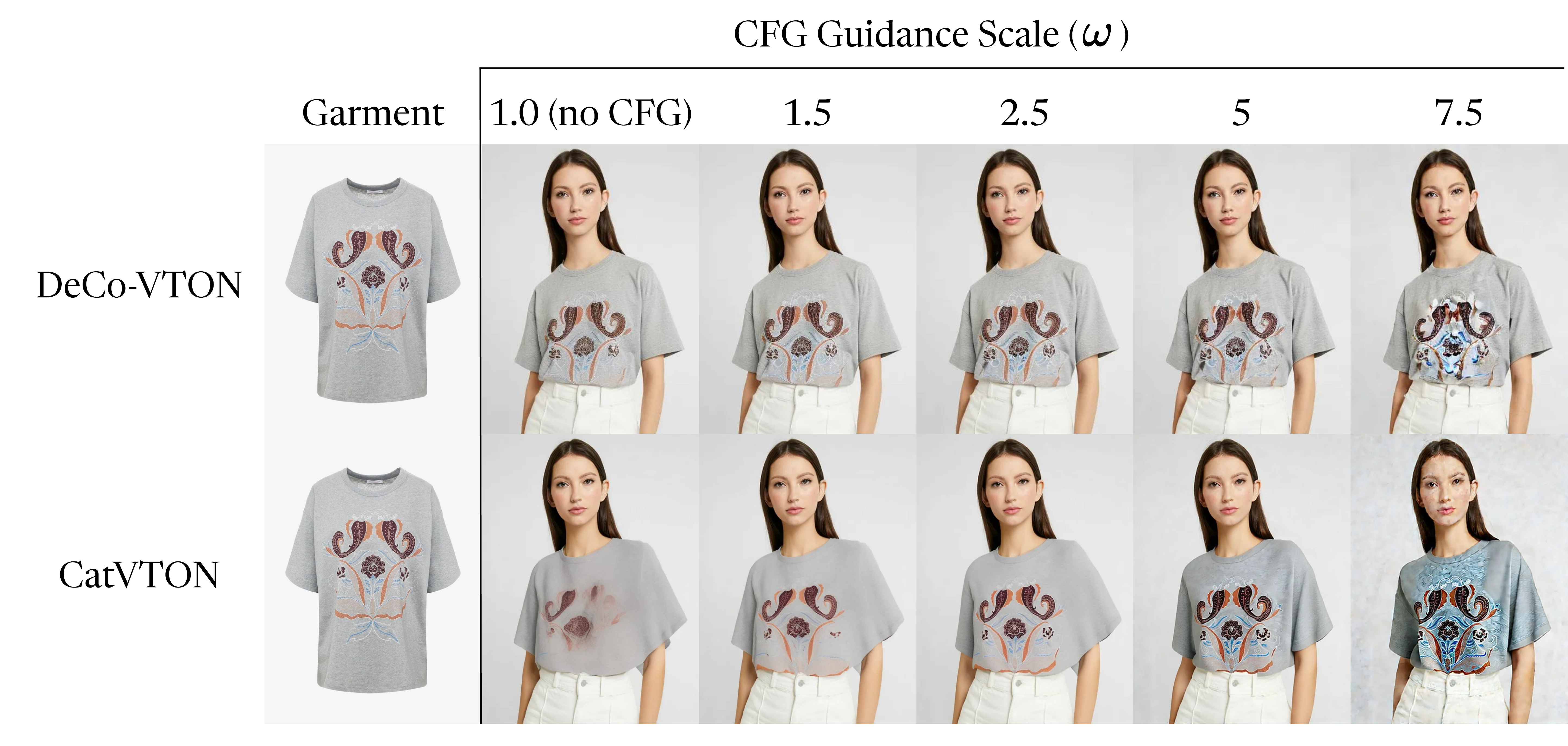}
  \caption{Effect of guidance scale~$\omega$. \textbf{Top:} DeCo-VTON preserves garment details even at $\omega{=}1.0$ and remains stable across scales. \textbf{Bottom:} CatVTON produces blurred garments at $\omega{=}1.0$ (Conflicts~2\&3) and introduces artifacts at higher~$\omega$ (Conflict~1).}
  \label{fig:cfg_analysis}
\end{figure}

Fig.~\ref{fig:cfg_analysis} provides the diagnostic evidence referenced in Sec.~\ref{sec:mismatch}.
At $\omega{=}1.0$ (no guidance), CatVTON already produces blurred garment details, confirming that Conflicts~2 and~3 degrade the conditional branch itself---the network's garment encoding capacity is compromised even before CFG is applied.
Increasing~$\omega$ further degrades CatVTON's output with artifacts and over-sharpening, confirming Conflict~1: the unconditional branch retains garment information, causing guidance to suppress rather than enhance garment details.

In contrast, DeCo-VTON preserves readable logos at $\omega{=}1.0$ and remains visually stable across scales, validating that the three conflicts have been resolved. Fig.~\ref{fig:fid_omega} quantitatively confirms this trend: CatVTON's FID degrades sharply beyond $\omega{=}2.5$, whereas DeCo-VTON remains stable across all tested scales.

\subsection{User Study}
\label{sec:user_study}

\begin{figure}[t]
  \begin{minipage}[t]{0.48\linewidth}
    \centering
    \includegraphics[width=\linewidth]{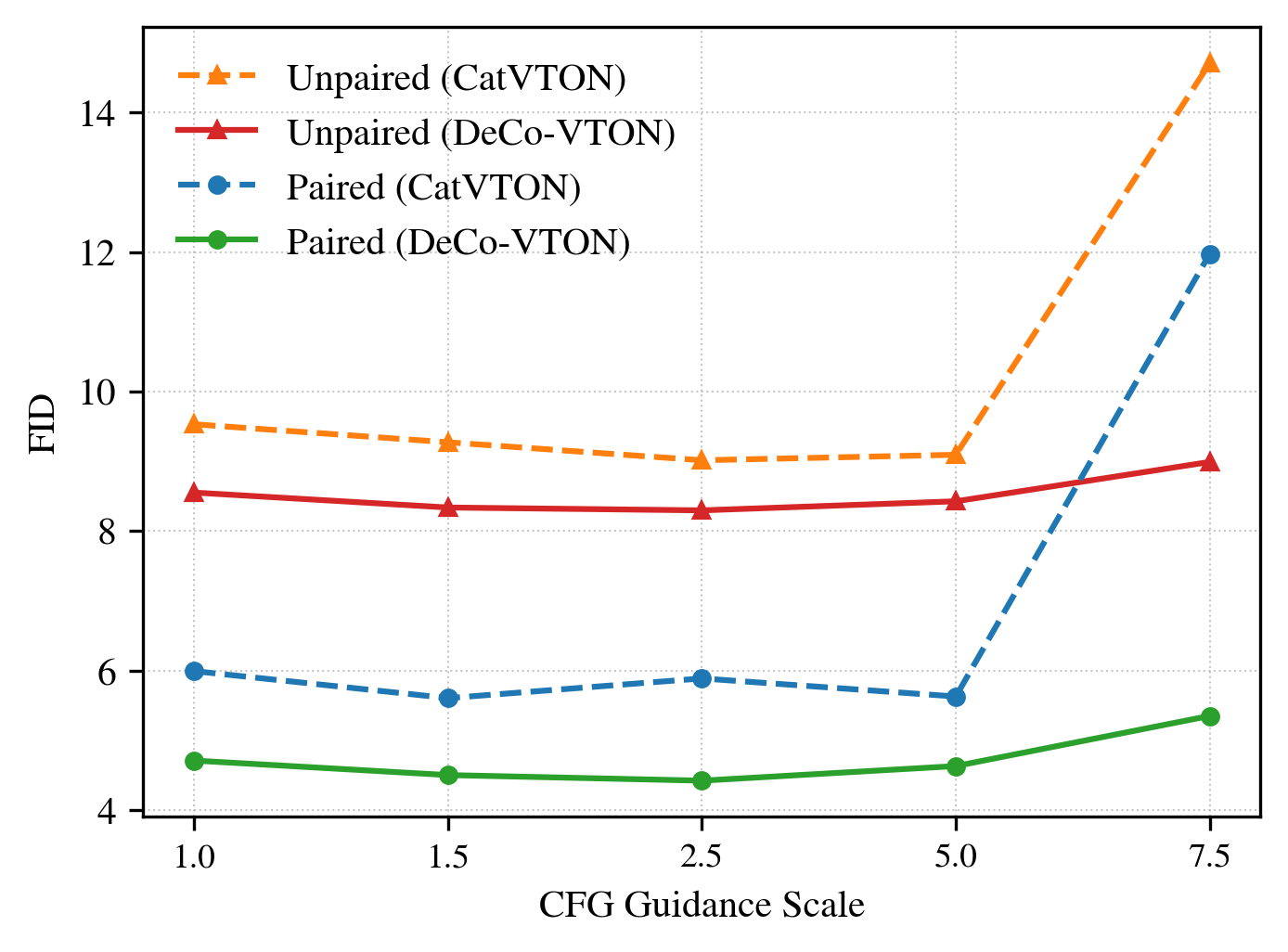}
    \caption{FID vs.\ guidance scale~$\omega$ on VITON-HD. 
    CatVTON degrades sharply at higher~$\omega$; DeCo-VTON 
    remains stable.}
    \label{fig:fid_omega}
  \end{minipage}
  \hfill
  \begin{minipage}[t]{0.48\linewidth}
    \centering
    \includegraphics[width=\linewidth]{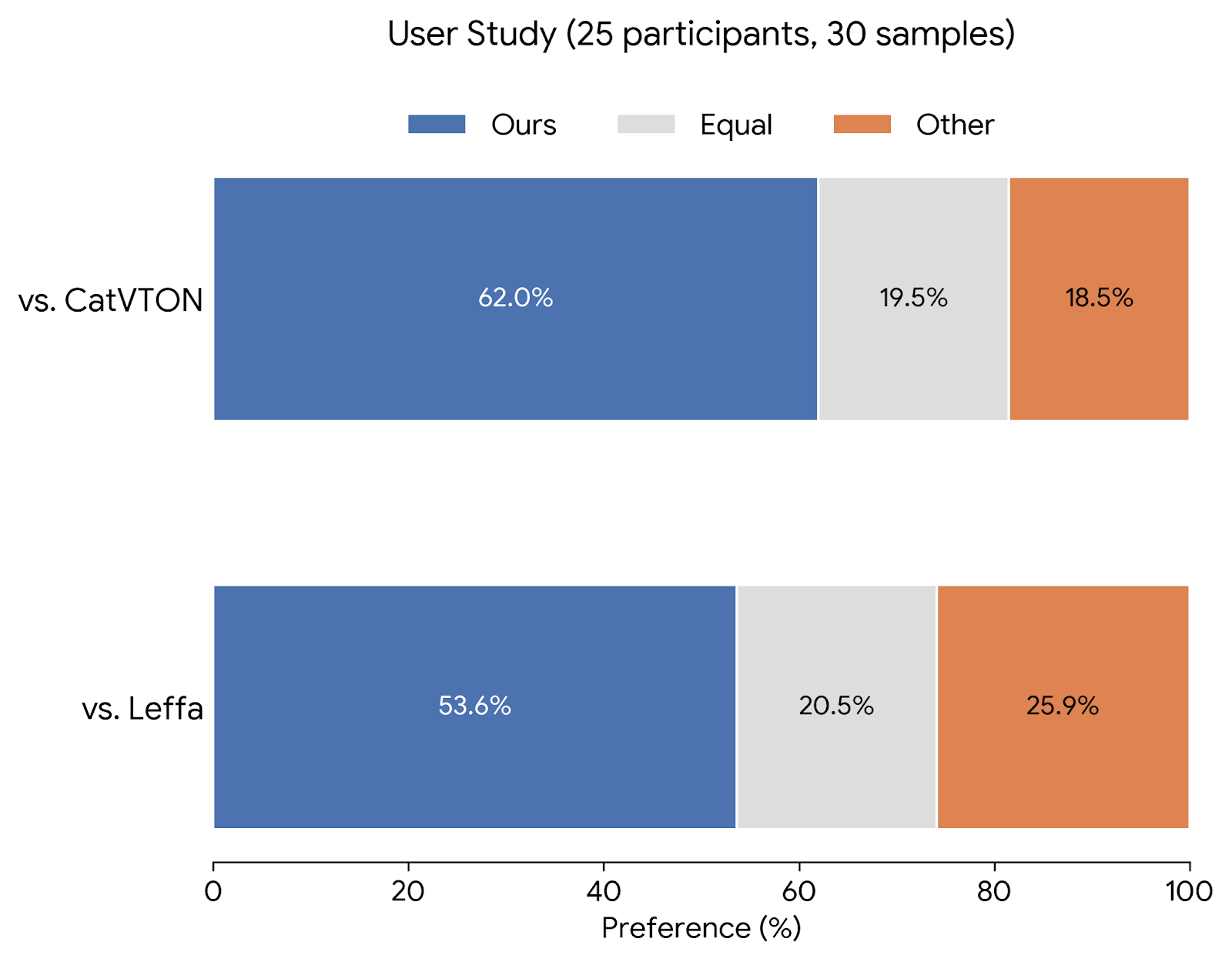}
    \caption{User study (25 participants, 30 samples; both $p{<}0.001$). DeCo-VTON is preferred over CatVTON at approximately 3.5:1 and over Leffa at approximately 2:1.}
    \label{fig:user_study}
  \end{minipage}%
\end{figure}

Fig.~\ref{fig:user_study} reports a randomized blind user study.
DeCo-VTON is preferred over CatVTON at approximately 3.5:1 and over Leffa at approximately 2:1, confirming that the quantitative gains translate to perceptually meaningful improvements.

\section{Conclusion}
\label{sec:conclusion}

This paper asked why full fine-tuning fails in spatial-concatenation-based virtual try-on and showed that the answer lies in a single insight: \emph{garment conditioning must be decoupled from the denoising process.}
By analyzing dual-UNet reference networks through noise prediction visualization, we identified that spatial concatenation violates this insight by embedding the garment within the denoising target, giving rise to three functional conflicts---guidance leakage, gradient competition, and train-test discrepancy.
Three design principles (Garment-Free Guidance, Decoupled Loss, and Clean Latent Anchoring) restore the necessary decoupling, thereby unlocking effective full fine-tuning for the first time under spatial concatenation.

The resulting model, DeCo-VTON, achieves state-of-the-art results among single-network methods and matches the dual-UNet state of the art at half the parameters, with no architectural modification.
This demonstrates that a properly designed conditioning recipe alone can unlock the potential of an existing backbone---the architecture was never the bottleneck; the unresolved conflict between conditioning and denoising was.

\noindent\textbf{Limitations and future work.}
Our principles are validated on UNet-based spatial concatenation with classifier-free guidance.
Flow-matching architectures such as Voost~\cite{voost} exhibit the same full fine-tuning failure (Sec.~\ref{sec:mismatch}), suggesting the mismatch is architecture-agnostic.

Among our three principles, only Garment-Free Guidance is tied to the CFG mechanism; Decoupled Loss and Clean Latent Anchoring instead address gradient competition and the train-test discrepancy, which arise from spatial concatenation itself and are thus independent of the guidance scheme, so we expect them to transfer to DiT and flow-matching backbones with only minor adaptation. Adapting GFG to non-CFG architectures would instead require an alternative mechanism that prevents garment information from influencing the baseline prediction without relying on a conditional-unconditional split.

More broadly, the conditioning--denoising conflict we identify is not specific to virtual try-on: any spatial-concatenation task in which a known signal is embedded within the denoising target, such as reference-based inpainting or outpainting, may benefit from the same decoupling principles.

\section*{Acknowledgement}
This research was supported by the Regional Innovation System \& Education (RISE) program through the Gyeongbuk RISE Center, funded by the Ministry of Education (MOE) and the Gyeongsangbuk-do, Republic of Korea. (2026-RISE-15-119)

\section*{Disclosure of Interests}
The authors have no competing interests to declare that are relevant to the content of this article.

%
%
\bibliographystyle{splncs04}
\bibliography{main}

\clearpage
\appendix

\begin{center}
  {\Large\bfseries DeCo-VTON: Supplementary Material\par}
\end{center}

\section{Experimental Details}
\label{sec:details}

\subsection{Hardware and Software Environment}
\label{sec:hardware}

All experiments were conducted on a Linux server equipped with an Intel Xeon Platinum 8570 CPU and two NVIDIA H200 (141\,GB) GPUs.
Our framework is implemented in Python 3.12 and PyTorch 2.8 with CUDA 12.9 support, built upon the \texttt{diffusers} and \texttt{accelerate} libraries from Hugging Face.
For fair comparison and reproducibility, we follow the evaluation metrics and overall implementation structure of the official CatVTON~\cite{catvton} repository.

\subsection{Training Configuration}
\label{sec:training_config}

We follow the same training configuration as CatVTON~\cite{catvton} with full fine-tuning enabled.
Table~\ref{tab:config} summarizes the complete hyperparameters.
We employ mixed-precision training (BF16) and keep the VAE encoder frozen throughout all experiments.
The noise scheduler follows the standard linear schedule of Stable Diffusion v1.5~\cite{ldm}.

\begin{table}[ht]
  \centering
  \footnotesize
  \caption{Detailed training hyperparameters.}
  \label{tab:config}
  \begin{tabular}{lc}
    \toprule
    Configuration & Value \\
    \midrule
    Base Model & Stable Diffusion v1.5 (Inpainting) \\
    Resolution & $512 \times 384$ \\
    Optimizer & AdamW \\
    Learning Rate & $1.0 \times 10^{-5}$ (Constant) \\
    Weight Decay & $1.0 \times 10^{-2}$ \\
    Adam $\beta_1, \beta_2$ & 0.9, 0.999 \\
    Gradient Clipping Norm & 1.0 \\
    Gradient Accumulation & 2 \\
    Global Batch Size & 128 \\
    Mixed Precision & BF16 \\
    Noise Scheduler & Linear \\
    Prediction Type & $\epsilon$-prediction \\
    Training Steps & 16k (VITON-HD) / 32k (DressCode) \\
    DREAM $\lambda$ & 10.0 \\
    CFG Dropout Probability & 0.1 \\
    \bottomrule
  \end{tabular}
\end{table}

\subsection{Details of the Decoupled Loss}
\label{sec:loss_details}

As discussed in Sec.~3.2 of the main paper, treating the garment region as a denoising target introduces gradient competition (Conflict~2).
To address this, we implement the Decoupled Loss inspired by the DREAM~\cite{dream} objective, applying rectification exclusively to the person region.

We first establish notation consistent with the main paper.
Let $\boldsymbol{\epsilon}_\theta(\cdot)$ denote the Denoising UNet, and let $\mathbf{X}_t$ denote the full input constructed via Eq.~1 of the main paper.
Given $\mathbf{X}_t$ at timestep $t$, the network predicts a noise tensor $\boldsymbol{\epsilon}_\theta = \boldsymbol{\epsilon}_\theta(\mathbf{X}_t, t)$, and $\bar{\boldsymbol{\epsilon}}_t$ denotes the ground-truth noise.
Since the person and garment latents are spatially concatenated along the height dimension, the predicted noise and the ground-truth noise naturally decompose into the person and garment regions:
\begin{equation}
  \boldsymbol{\epsilon}_\theta = \boldsymbol{\epsilon}^p_\theta \oplus \boldsymbol{\epsilon}^g_\theta,
  \quad
  \bar{\boldsymbol{\epsilon}}_t = \bar{\boldsymbol{\epsilon}}^p_t \oplus \bar{\boldsymbol{\epsilon}}^g_t,
\end{equation}
where $\oplus$ denotes spatial concatenation along the height dimension (as in the main paper), and superscripts $p$ and $g$ refer to the person and garment regions, respectively.

We apply the DREAM-style rectification only to the person region, which requires rectifying both the regression target and the input latent.
First, the rectified target for the person region, $\bar{\boldsymbol{\epsilon}}^p_{\text{dream},t}$, is defined as
\begin{equation}
  \bar{\boldsymbol{\epsilon}}^p_{\text{dream},t}
  =
  \bar{\boldsymbol{\epsilon}}^p_t
  +
  \omega_t(\lambda)\,
  \bigl(
    \bar{\boldsymbol{\epsilon}}^p_t - \boldsymbol{\epsilon}^p_{\theta,\text{sg}}
  \bigr),
  \label{eq:dream_target}
\end{equation}
where (i) $\bar{\boldsymbol{\epsilon}}^p_t$ is the ground-truth noise added to the person latent, (ii) $\boldsymbol{\epsilon}^p_{\theta,\text{sg}}$ is the noise prediction on the person region from a frozen copy of the UNet (stop-gradient), and (iii) $\omega_t(\lambda) = (1 - \bar{\alpha}_t)^{\lambda / 2}$ is a time-dependent balancing weight.
Here, $\bar{\alpha}_t$ is the standard noise-schedule term, and $\lambda$ controls the strength of the rectification (we set $\lambda = 10$).

Next, we rectify the input latent.
Let $\bar{\mathbf{z}}_t = \bar{\mathbf{z}}^p_t \oplus \bar{\mathbf{z}}^g_t$ denote the ground-truth noisy latent obtained from the forward diffusion process.
We modify only the person region while keeping the garment region fixed:
\begin{equation}
  \tilde{\mathbf{z}}^p_t
  =
  \bar{\mathbf{z}}^p_t
  +
  \sqrt{1 - \bar{\alpha}_t}\,
  \omega_t(\lambda)\,
  \bigl(
    \bar{\boldsymbol{\epsilon}}^p_t - \boldsymbol{\epsilon}^p_{\theta,\text{sg}}
  \bigr),
  \quad
  \tilde{\mathbf{z}}^g_t = \bar{\mathbf{z}}^g_t,
  \label{eq:input_rectification}
\end{equation}
and we define $\tilde{\mathbf{z}}_t = \tilde{\mathbf{z}}^p_t \oplus \tilde{\mathbf{z}}^g_t$ as the rectified noisy latent at timestep $t$.
The rectified input $\tilde{\mathbf{X}}_t$ is then constructed by substituting $\tilde{\mathbf{z}}_t$ into Eq.~1 of the main paper while leaving all other components (mask and clean latents) unchanged.
Here, the bar ($\bar{\cdot}$) denotes the ground-truth noisy latent from the forward diffusion process, and the tilde ($\tilde{\cdot}$) denotes the rectified latent used for DREAM-style training.
Note that keeping the garment latent at its forward-diffused value ($\tilde{\mathbf{z}}^g_t = \bar{\mathbf{z}}^g_t$) during training mirrors the role that Clean Latent Anchoring (Sec.~4.4 of the main paper) plays at inference: both ensure that the garment region follows the correct noise schedule rather than being corrupted by prediction errors.

Finally, we minimize the reconstruction error only on the person region using the rectified input and target.
We denote by $\boldsymbol{\epsilon}^p_\theta(\tilde{\mathbf{X}}_t, t)$ the person-region portion of the noise prediction given the rectified input.
The resulting loss is
\begin{equation}
  \mathcal{L}_{\text{decoupled}}
  =
  \mathbb{E}_{\mathbf{z}_0, \boldsymbol{\epsilon}, t}
  \Bigl[
    \bigl\|
      \bar{\boldsymbol{\epsilon}}^p_{\text{dream},t}
      -
      \boldsymbol{\epsilon}^p_\theta(\tilde{\mathbf{X}}_t, t)
    \bigr\|^2
  \Bigr].
  \label{eq:decoupled_loss_supp}
\end{equation}
By computing the loss exclusively on the person region, the network is never trained to reconstruct the garment---the garment region serves purely as conditioning context, analogous to the reconstruction-free reference UNet in dual-UNet methods.

\subsection{Efficiency Measurement Details}
\label{sec:efficiency_details}

We describe the measurement protocol for Table~3 of the main paper.
For model size and FLOPs, we measure the number of parameters and total GFLOPs per $512{\times}384$ image using a FLOPs profiler based on \texttt{calflops}, applied to all modules except the VAE decoder, which is nearly identical across methods.
For inference efficiency, we run all models on a single NVIDIA H200 GPU with batch size 1 using FP16 (PyTorch AMP).
We perform 5 warm-up runs and then measure the average latency over 30 consecutive inference runs, while recording the peak GPU memory using \texttt{torch.cuda.max\_memory\_allocated}.

\subsection{User Study Protocol}
\label{sec:user_study_protocol}

We describe the protocol for the user study reported in Fig.~7 of the main paper.
We conducted a randomized blind study with 25 participants evaluating 30 samples across two comparison pairs (DeCo-VTON vs.\ CatVTON~\cite{catvton} and DeCo-VTON vs.\ Leffa~\cite{leffa}), yielding a total of 1,500 responses.
Each trial presented two images in A/B format with an ``Equal'' option, where participants selected the preferred result or indicated no preference.
Both the sample order and the A/B placement were fully randomized per participant to eliminate ordering bias.
Statistical significance was verified using both binomial and chi-squared tests, with both confirming $p < 0.001$ for each comparison pair.

\section{Cross-Architecture Comparison with Spatial-Concatenation DiT}
\label{sec:voost}

The main paper identifies spatial concatenation's structural limitation---the loss of decoupling between garment conditioning and denoising---as architecture-agnostic (Sec.~3.2 of the main paper).
We validate this claim by comparing with Voost~\cite{voost}, an 11.9B-parameter DiT that adopts the same spatial-concatenation conditioning as CatVTON~\cite{catvton}, placing the garment and person images side by side and denoising the composite as a whole.

\noindent\textbf{Full fine-tuning failure in Voost.}
Voost reports that full fine-tuning (11.9B parameters) yields worse performance than attention-only training (2.69B parameters) on VITON-HD: FID~6.351 vs.\ 5.269, KID~0.886 vs.\ 0.404 (Table~3 of \cite{voost}).
This directly mirrors the pattern analyzed in the main paper: under spatial concatenation, expanding the set of trainable parameters without resolving the conditioning--denoising conflict degrades rather than improves generation quality.
Voost attributes this to ``overfitting''; our analysis provides a more structural explanation through the three identified conflicts.

Note that Voost employs flow matching without classifier-free guidance, meaning Conflict~1 (guidance leakage) does not apply.
However, Conflicts~2 (gradient competition) and~3 (train-test discrepancy) arise from spatial concatenation itself, independent of the guidance mechanism, and are consistent with the observed failure.

\noindent\textbf{Quantitative comparison at \boldmath$1024{\times}768$.}
Tables~\ref{tab:voost_vitonhd} and~\ref{tab:voost_dresscode} report results at Voost's native resolution on VITON-HD and DressCode, respectively.
DeCo-VTON is evaluated at matched resolution.

\begin{table}[t]
  \centering
  \footnotesize
  \setlength{\tabcolsep}{4pt}
  \caption{Comparison on VITON-HD at $1024{\times}768$. All baseline results from Voost~\cite{voost}. DeCo-VTON is trained at this $1024\times768$ resolution to ensure a fair spatial comparison. \textbf{Best} in bold, \underline{second-best} underlined.}
  \label{tab:voost_vitonhd}
  \begin{tabular}{l c cccc cc}
    \toprule
    \multirow{2}{*}{Method} & \multirow{2}{*}{Params} & \multicolumn{4}{c}{Paired} & \multicolumn{2}{c}{Unpaired} \\
    \cmidrule(lr){3-6} \cmidrule(lr){7-8}
    & & FID$\downarrow$ & KID$\downarrow$ & SSIM$\uparrow$ & LPIPS$\downarrow$ & FID$\downarrow$ & KID$\downarrow$ \\
    \midrule
    OOTDiffusion~\cite{ootd} & 2.2B & 6.520 & 0.896 & 0.851 & 0.096 & 9.672 & 1.206 \\
    IDM-VTON~\cite{idm} & 7.0B & 6.343 & 1.322 & 0.881 & 0.079 & 9.613 & 1.639 \\
    CatVTON~\cite{catvton} & 860M & 6.141 & 0.964 & 0.869 & 0.097 & 9.141 & 1.267 \\
    Leffa~\cite{leffa} & 1.8B & 6.310 & 1.208 & 0.872 & 0.081 & 9.442 & 1.249 \\
    Voost (VTON-only)~\cite{voost} & 11.9B & 5.804 & 0.618 & 0.868 & 0.079 & 9.112 & 1.051 \\
    Voost~\cite{voost} & 11.9B & \underline{5.269} & \textbf{0.404} & \textbf{0.898} & \textbf{0.056} & \underline{8.982} & \underline{0.899} \\
    \midrule
    \textbf{DeCo-VTON (Ours)} & 860M & \textbf{4.741} & \underline{0.429} & \underline{0.888} & \underline{0.067} & \textbf{8.675} & \textbf{0.832} \\
    \bottomrule
  \end{tabular}
\end{table}

\begin{table}[t]
  \centering
  \footnotesize
  \setlength{\tabcolsep}{4pt}
  \caption{Comparison on DressCode at $1024{\times}768$. DeCo-VTON is trained at this $1024\times768$ resolution to ensure a fair spatial comparison. Notation follows Table~\ref{tab:voost_vitonhd}.}
  \label{tab:voost_dresscode}
  \begin{tabular}{l c cccc cc}
    \toprule
    \multirow{2}{*}{Method} & \multirow{2}{*}{Params} & \multicolumn{4}{c}{Paired} & \multicolumn{2}{c}{Unpaired} \\
    \cmidrule(lr){3-6} \cmidrule(lr){7-8}
    & & FID$\downarrow$ & KID$\downarrow$ & SSIM$\uparrow$ & LPIPS$\downarrow$ & FID$\downarrow$ & KID$\downarrow$ \\
    \midrule
    OOTDiffusion~\cite{ootd} & 2.2B & 3.953 & 0.720 & 0.898 & 0.073 & 6.704 & 1.863 \\
    IDM-VTON~\cite{idm} & 7.0B & 3.801 & 1.201 & \underline{0.923} & \underline{0.048} & 5.621 & 1.554 \\
    CatVTON~\cite{catvton} & 860M & 3.283 & 0.670 & 0.901 & 0.071 & 5.424 & 1.549 \\
    Leffa~\cite{leffa} & 1.8B & 3.651 & 0.709 & 0.911 & 0.060 & 5.462 & 1.528 \\
    Voost (VTON-only)~\cite{voost} & 11.9B & 3.043 & 0.565 & 0.910 & 0.052 & 5.298 & 1.132 \\
    Voost~\cite{voost} & 11.9B & \underline{2.787} & \underline{0.377} & \textbf{0.933} & \textbf{0.044} & \underline{5.081} & \textbf{0.787} \\
    \midrule
    \textbf{DeCo-VTON (Ours)} & 860M & \textbf{2.216} & \textbf{0.141} & 0.911 & 0.051 & \textbf{4.650} & \underline{0.795} \\
    \bottomrule
  \end{tabular}
\end{table}

Despite having 1/14th the total parameters (860M vs.\ 11.9B), DeCo-VTON achieves the best paired FID on both datasets and the best unpaired FID and KID on VITON-HD, outperforming Voost (VTON-only) across all FID and KID metrics.
Even compared to the full Voost---which additionally benefits from joint VTON+VTOFF training, an orthogonal contribution---DeCo-VTON achieves lower FID on both datasets in both paired and unpaired settings, and lower paired KID on DressCode.
The full Voost retains an advantage in SSIM and LPIPS, likely reflecting the additional supervision from the VTOFF auxiliary task.
These results confirm that the conditioning--denoising conflict, not the backbone capacity, is the primary bottleneck under spatial concatenation.

\section{Comparison with Sequence-Concatenation DiT}
\label{sec:refton}

While our analysis and design principles target spatial concatenation specifically, we additionally compare with RefTON~\cite{refton}, a Flux-Kontext-based DiT model that employs a different conditioning paradigm.
RefTON encodes each input (person, garment, optional reference) independently and concatenates the resulting token sequences along the sequence dimension, with a position index distinguishing different conditions.
Unlike spatial concatenation, the garment is not embedded within the denoising target; instead, it is processed as a separate token group.
We include this comparison to contextualize DeCo-VTON's performance across conditioning paradigms, rather than to validate the decoupling principles.

Tables~\ref{tab:refton_vitonhd} and~\ref{tab:refton_dresscode} report results at $512{\times}384$ on VITON-HD and DressCode, respectively.
RefTON is based on Flux-Kontext (12B+ backbone) with LoRA fine-tuning; ``+R'' denotes the use of an additional reference image showing the garment worn by a different person, an input not available to other methods.
Note that RefTON trains at $1024{\times}768$ and resizes to $512{\times}384$ for evaluation, whereas DeCo-VTON trains and evaluates at $512{\times}384$.

\begin{table}[t]
  \centering
  \footnotesize
  \setlength{\tabcolsep}{4pt}
  \caption{Comparison with RefTON on VITON-HD at $512{\times}384$. All RefTON results from~\cite{refton}. \textbf{Best} in bold.}
  \label{tab:refton_vitonhd}
  \begin{tabular}{l cccc cc}
    \toprule
    \multirow{2}{*}{Method} & \multicolumn{4}{c}{Paired} & \multicolumn{2}{c}{Unpaired} \\
    \cmidrule(lr){2-5} \cmidrule(lr){6-7}
    & FID$\downarrow$ & KID$\downarrow$ & SSIM$\uparrow$ & LPIPS$\downarrow$ & FID$\downarrow$ & KID$\downarrow$ \\
    \midrule
    RefTON~\cite{refton}    & 5.45 & 0.82 & 0.873 & 0.057 & 8.58 & 1.06 \\
    RefTON+R~\cite{refton}  & 4.69 & 0.68 & 0.879 & 0.049 & 8.43 & 0.91 \\
    \midrule
    \textbf{DeCo-VTON (Ours)} & \textbf{4.438} & \textbf{0.010} & \textbf{0.880} & \textbf{0.047} & \textbf{8.266} & \textbf{0.517} \\
    \bottomrule
  \end{tabular}
\end{table}

\begin{table}[t]
  \centering
  \footnotesize
  \setlength{\tabcolsep}{4pt}
  \caption{Comparison with RefTON on DressCode at $512{\times}384$. Notation follows Table~\ref{tab:refton_vitonhd}.}
  \label{tab:refton_dresscode}
  \begin{tabular}{l cccc cc}
    \toprule
    \multirow{2}{*}{Method} & \multicolumn{4}{c}{Paired} & \multicolumn{2}{c}{Unpaired} \\
    \cmidrule(lr){2-5} \cmidrule(lr){6-7}
    & FID$\downarrow$ & KID$\downarrow$ & SSIM$\uparrow$ & LPIPS$\downarrow$ & FID$\downarrow$ & KID$\downarrow$ \\
    \midrule
    RefTON~\cite{refton}    & 3.48 & 1.20 & 0.912 & 0.037 & 5.31 & 1.36 \\
    RefTON+R~\cite{refton}  & 2.94 & 0.95 & \textbf{0.918} & \textbf{0.031} & 5.07 & 1.15 \\
    \midrule
    \textbf{DeCo-VTON (Ours)} & \textbf{2.175} & \textbf{0.062} & 0.914 & \textbf{0.031} & \textbf{4.310} & \textbf{0.628} \\
    \bottomrule
  \end{tabular}
\end{table}

Under identical input conditions (garment and masked person only), DeCo-VTON outperforms RefTON across all metrics on both VITON-HD and DressCode.
Notably, DeCo-VTON surpasses even RefTON+R---which receives an additional reference image showing the garment worn by a different person---in FID and KID on both datasets, while matching or exceeding LPIPS.
These results are achieved with an 860M-parameter UNet, compared to RefTON's 12B+ Flux-Kontext backbone, suggesting that DeCo-VTON's conditioning recipe generalizes competitively even against models with fundamentally different conditioning paradigms and over an order of magnitude more parameters.

\section{Additional Results}
\label{sec:additional}

\subsection{Leave-one-out Analysis of the Decoupled Loss}
\label{sec:dl_ablation}
Table~4 of the main paper validates each design principle through progressive addition (Full FT $\to$ +GFG $\to$ +DL $\to$ +CLA). To isolate the contribution of the Decoupled Loss (DL) independent of ablation ordering, Table~\ref{tab:dl_ablation} compares the full recipe (+GFG+DL+CLA) against a variant that omits only DL (+GFG+CLA) at both $512{\times}384$ and $1024{\times}768$ resolutions on VITON-HD.

\begin{table}[ht]
  \centering
  \footnotesize
  \setlength{\tabcolsep}{2pt}
  \caption{Leave-one-out analysis of the Decoupled Loss on VITON-HD. \textbf{Best} in bold.}
  \label{tab:dl_ablation}
  \begin{tabular}{l l cccc cc}
    \toprule
    \multirow{2}{*}{Resolution}
      & \multirow{2}{*}{Configuration}
      & \multicolumn{4}{c}{Paired}
      & \multicolumn{2}{c}{Unpaired} \\
    \cmidrule(lr){3-6} \cmidrule(lr){7-8}
    & & FID$\downarrow$ & KID$\downarrow$ & SSIM$\uparrow$ & LPIPS$\downarrow$ & FID$\downarrow$ & KID$\downarrow$ \\
    \midrule
    \multirow{3}{*}{$512{\times}384$}
    & Full FT (baseline) & 5.250 & 0.402 & 0.869 & 0.055 & 8.813 & 0.956 \\
    & + GFG + CLA (no DL) & 4.449 & \textbf{0.004} & 0.880 & 0.047 & 8.308 & 0.528 \\
    & + GFG + DL + CLA (Ours) & \textbf{4.438} & 0.010 & \textbf{0.880} & \textbf{0.047} & \textbf{8.266} & \textbf{0.517} \\
    \midrule
    \multirow{2}{*}{$1024{\times}768$}
    & + GFG + CLA (no DL) & 4.919 & \textbf{0.422} & 0.884 & 0.070 & 8.988 & 0.981 \\
    & + GFG + DL + CLA (Ours) & \textbf{4.741} & 0.429 & \textbf{0.888} & \textbf{0.067} & \textbf{8.675} & \textbf{0.832} \\
    \bottomrule
  \end{tabular}
\end{table}

At $512{\times}384$, GFG and CLA account for the primary gains; DL adds modest improvement on unpaired metrics (FID~$8.308{\to}8.266$, KID~$0.528{\to}0.517$), consistent with Table~4 of the main paper regardless of ablation ordering. At $1024{\times}768$, however, DL's contribution becomes more pronounced: removing DL degrades unpaired FID by $0.313$ and KID by $0.149$.
This resolution-dependent behavior suggests that DL's role in eliminating gradient competition grows with the distributional complexity of higher resolutions, where the garment reconstruction and person generation objectives diverge more substantially.

\subsection{Per-category Analysis on DressCode}
\label{sec:percategory}

To verify that our improvements generalize across garment types, Table~\ref{tab:percategory} reports per-category results on DressCode.
DeCo-VTON consistently outperforms CatVTON across upper body, lower body, and dresses in FID, KID, and LPIPS under both paired and unpaired settings, while also achieving comparable or higher SSIM.

\begin{table}[ht]
  \centering
  \footnotesize
  \setlength{\tabcolsep}{4pt}
  \caption{Per-category comparison on DressCode. \textbf{Best} in bold.}
  \label{tab:percategory}
  \begin{tabular}{l cccc cc}
    \toprule
    \multirow{2}{*}{Method} & \multicolumn{4}{c}{Paired} & \multicolumn{2}{c}{Unpaired} \\
    \cmidrule(lr){2-5} \cmidrule(lr){6-7}
    & FID$\downarrow$ & KID$\downarrow$ & SSIM$\uparrow$ & LPIPS$\downarrow$ & FID$\downarrow$ & KID$\downarrow$ \\
    \midrule
    \multicolumn{7}{c}{\textit{Upper body}} \\
    CatVTON~\cite{catvton} & 7.40 & 0.498 & 0.933 & 0.031 & 11.81 & 1.431 \\
    \textbf{DeCo-VTON (Ours)} & \textbf{6.17} & \textbf{0.102} & \textbf{0.939} & \textbf{0.024} & \textbf{10.82} & \textbf{0.720} \\
    \midrule
    \multicolumn{7}{c}{\textit{Lower body}} \\
    CatVTON~\cite{catvton} & 8.30 & 1.104 & 0.920 & 0.037 & 13.52 & 2.741 \\
    \textbf{DeCo-VTON (Ours)} & \textbf{6.21} & \textbf{0.196} & \textbf{0.926} & \textbf{0.029} & \textbf{12.11} & \textbf{1.375} \\
    \midrule
    \multicolumn{7}{c}{\textit{Dresses}} \\
    CatVTON~\cite{catvton} & 7.66 & 0.487 & 0.862 & 0.055 & 10.90 & 1.147 \\
    \textbf{DeCo-VTON (Ours)} & \textbf{6.45} & \textbf{0.058} & \textbf{0.872} & \textbf{0.043} & \textbf{10.63} & \textbf{0.909} \\
    \bottomrule
  \end{tabular}
\end{table}

\subsection{Additional Qualitative Results}
\label{sec:qual}

\begin{figure*}[t]
  \centering
  \includegraphics[height=0.9\textheight]{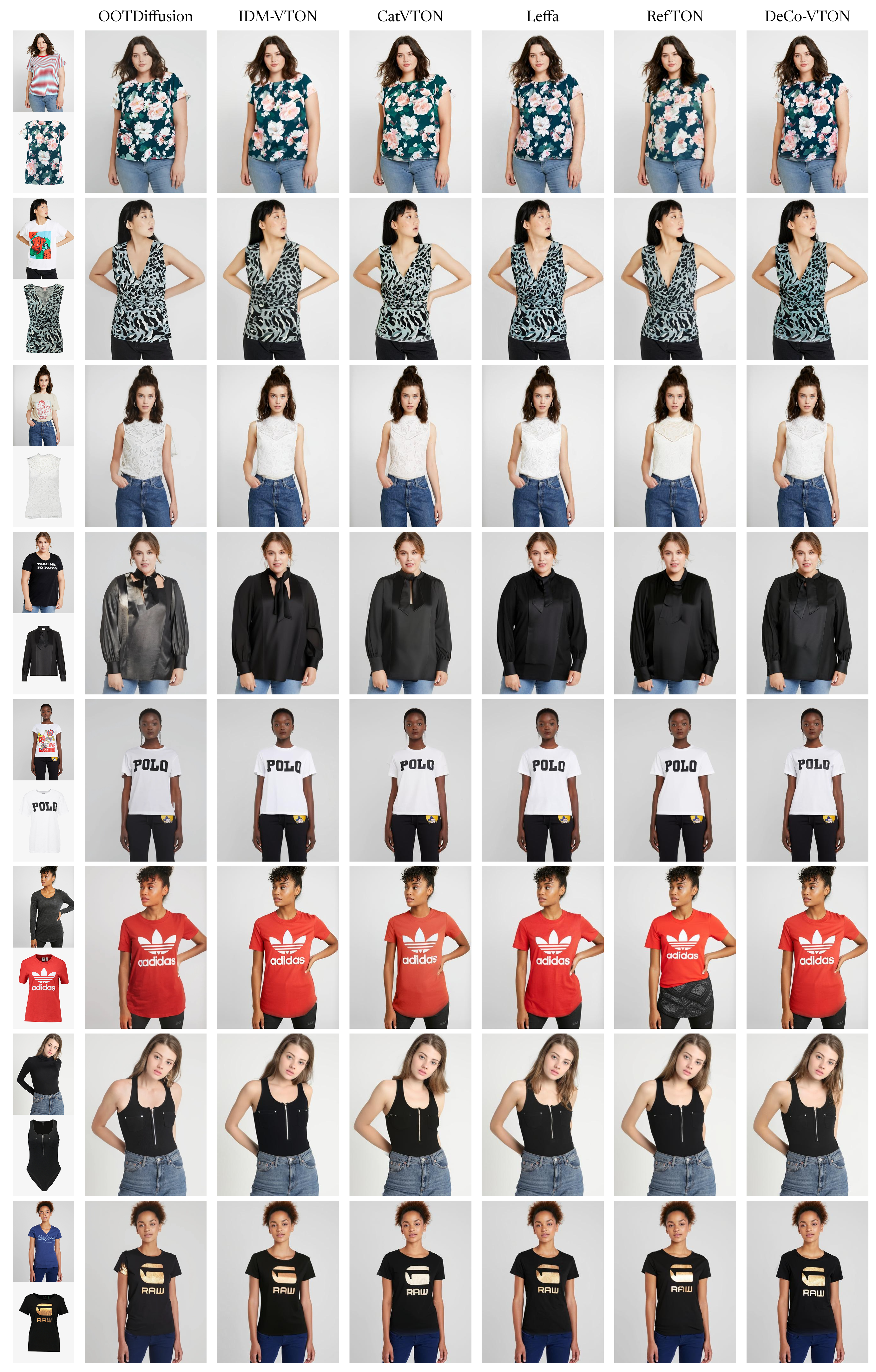}
  \caption{Additional qualitative comparisons on the VITON-HD dataset.
  For each example, we show the input person, the target garment, the ground-truth outfit, and the outputs of OOTDiffusion~\cite{ootd}, IDM-VTON~\cite{idm}, CatVTON~\cite{catvton}, Leffa~\cite{leffa}, RefTON~\cite{refton}, and DeCo-VTON (ours).}
  \label{fig:supp_vitonhd}
\end{figure*}

\begin{figure*}[t]
  \centering
  \includegraphics[height=0.9\textheight]{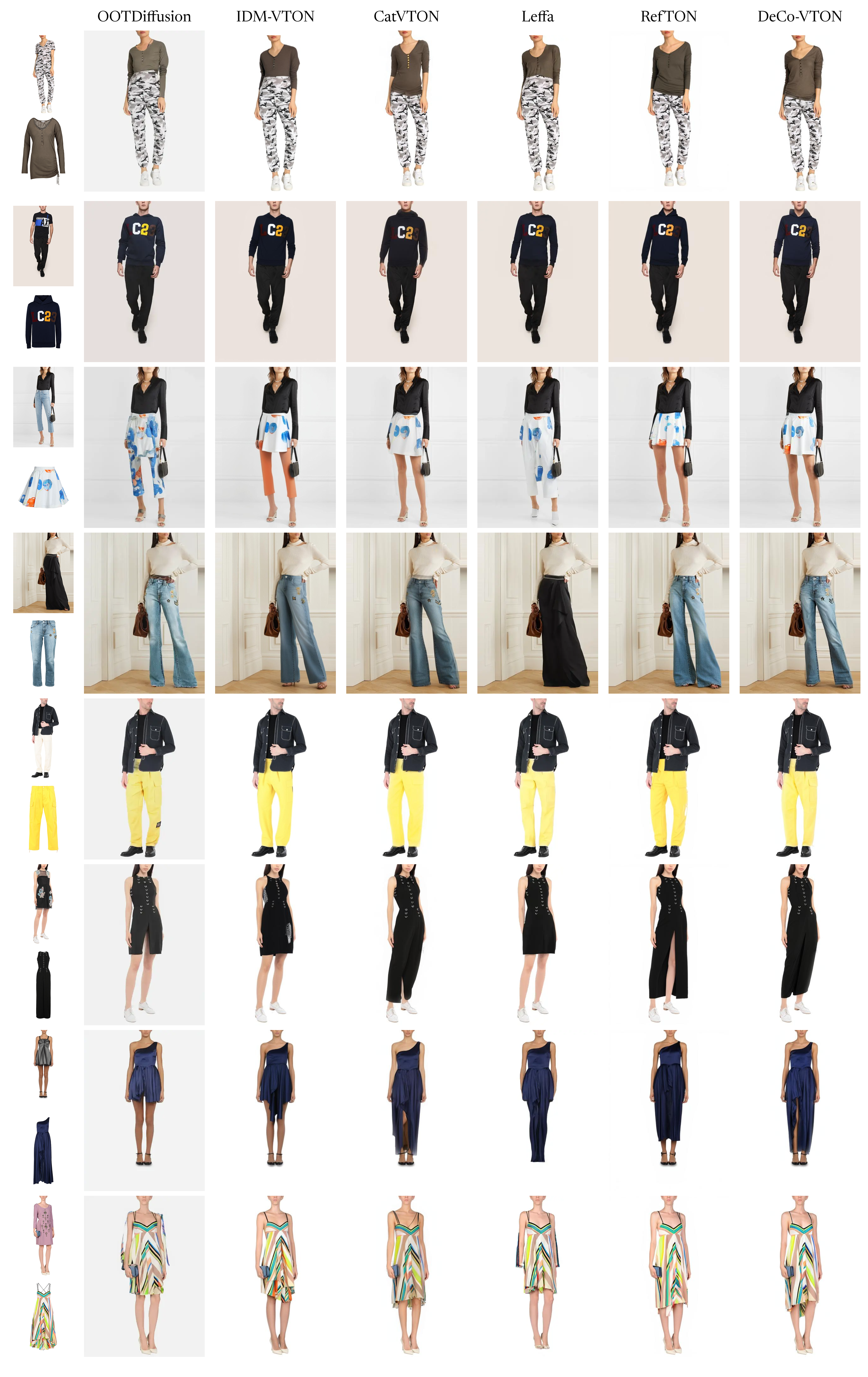}
  \caption{Additional qualitative comparisons on the DressCode dataset.
  We use the same ordering of inputs and methods as in Fig.~\ref{fig:supp_vitonhd}.}
  \label{fig:supp_dresscode}
\end{figure*}

We provide additional qualitative comparisons on VITON-HD and DressCode in Figs.~\ref{fig:supp_vitonhd} and~\ref{fig:supp_dresscode}.
The examples cover diverse garments and complex patterns.

\end{document}